\documentclass[11pt,a4paper]{article}
\usepackage{times}
\usepackage[utf8]{inputenc}
\usepackage{tcolorbox}
\usepackage{url}
\usepackage{authblk}
\usepackage[hyperref]{acl2018}
\usepackage{balance}

\title{On Generality and Knowledge Transferability in Cross-Domain Duplicate Question Detection for Heterogeneous Community Question Answering}

\aclfinalcopy 

\author[1]{Mohomed Shazan Mohomed Jabbar}
\author[1]{Luke Kumar}
\author[2]{Hamman Samuel}
\author[3]{Mi-Young Kim}
\author[1]{\\Sankalp Prabhakar}
\author[2]{Randy Goebel}
\author[2]{Osmar Zaïane}
\affil[1]{Alberta Machine Intelligence Institute (Amii), Edmonton, Canada}
\affil[2]{Department of Computing Science, University of Alberta, Edmonton, Canada}
\affil[3]{Department of Science, Augustana Faculty, University of Alberta, Camrose, Canada}
\affil[ ]{\fontsize{10pt}{12pt} \texttt{\{mohomedj,lkumar,hwsamuel,miyoung2,sankalp,rgoebel,zaiane\}@ualberta.ca}}

\date{}

\begin{document}
\maketitle
\begin{abstract}
Duplicate question detection is an ongoing challenge in community question answering because semantically equivalent questions can have significantly different words and structures. In addition, the identification of duplicate questions can reduce the resources required for retrieval, when the same questions are not repeated. This study compares the performance of deep neural networks and gradient tree boosting, and explores the possibility of domain adaptation with transfer learning to improve the under-performing target domains for the text-pair duplicates classification task, using three heterogeneous datasets: general-purpose Quora, technical Ask Ubuntu, and academic English Stack Exchange. Ultimately, our study exposes the alternative hypothesis that the meaning of a ``duplicate'' is not inherently general-purpose, but rather is dependent on the domain of learning, hence reducing the chance of transfer learning through adapting to the domain.
\end{abstract}

\section{Introduction}
\label{sec:intro}
To efficiently exploit Community Question Answering (CQA) forums, users need to know if their question has already been asked, to avoid re-posting a duplicate question. The identification of duplicate questions in CQA forums can provide at least three main advantages. Firstly, finding duplicate questions saves users' time because they do not have to wait for responses. Secondly, users searching for questions will be presented with better results with duplicates pruned. Thirdly, the overall retrievability of information for the CQA forum will be enhanced by reducing duplication.

Identifying two questions as duplicates can be challenging because the choice of words, structure of sentences, and even context, can vary significantly between questions, even if the intended semantics are near identical. 

\begin{figure}[ht!]
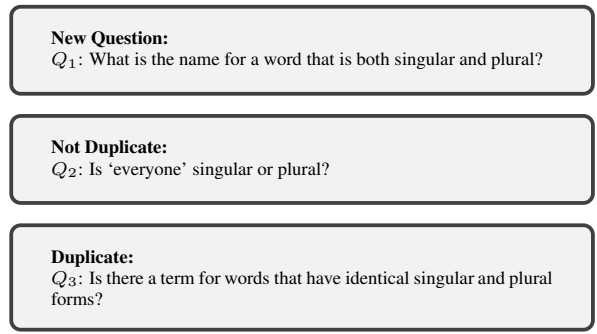

\scriptsize
	\begin{tcolorbox}
		\textbf{New Question:}\\
		$Q_1$: What is the name for a word that is both singular and plural?
	\end{tcolorbox}
	\begin{tcolorbox}
		\textbf{Not Duplicate:}\\
		$Q_2$: Is `everyone' singular or plural?
	\end{tcolorbox}
	\begin{tcolorbox}
		\textbf{Duplicate:}\\
		$Q_3$: Is there a term for words that have identical singular and plural forms?
	\end{tcolorbox}
\caption{Duplicate Question Detection Example}
\label{fig:1}
\end{figure}

In addition, questions with similar verbiage are not necessarily duplicates. Traditional computational information retrieval and Natural Language Processing (NLP) methods have achieved only limited success in detecting semantically identical text-pairs. Popular CQA forums like Quora and Stack Exchange (SE) have many new questions posted daily, some of which have been previously asked but have variations in wordings, synonyms, phrases, or sentence structure. Figure~\ref{fig:1} illustrates the practical difficulties of duplicate detection, given an incoming question, and potential matching existing questions.

When comparing state-of-the-art machine learning methods for this task, an interesting observation is that a classification model trained on a dataset from one domain cannot achieve the same performance to predict text-pair duplicates in another domain. For instance, the similarity between two question pairs can be completely different depending on the domain of the dataset which was used to train the classification model. The question pair ``Where can I find a place to eat pizza?'' and ``What's the closest Italian restaurant?'' can be classified as duplicate or not, depending on the domain of the dataset used in model training. With Quora, the similarity was 6\%, while with Stack Exchange, it was 46\%~\cite{Spacy2017}.

We present the results of an empirical analysis of popular state-of-the-art machine learning methods for text-pair duplicate classification: deep neural networks and gradient tree boosting, and explore the possibility of domain adaptation to increase the performance of under performing domains using transfer learning. 

We address three questions with our research goals. Firstly, we investigate the best approaches for text-pair duplicate detection. Secondly, we explore the possibility of a general-purpose cross-domain duplicate detection approach for heterogeneous datasets. Ultimately, we determine whether dataset domain affects the outcomes of the trained model, and evaluate the null hypothesis that the meaning of a ``duplicate'' is universal.

\section{Methodology}
\label{sec:method}
We used three publicly available datasets from \href{https://data.quora.com/First-Quora-Dataset-Release-Question-Pairs}{Quora}, and \href{https://archive.org/details/stackexchange}{Stack Exchange's Ask Ubuntu and English forums}. The datasets contain forum moderator annotated labels for duplicate and non-duplicate question pairs. Our study used only the question's title; the question's full body, tags and other meta data were not used for the SE datasets in order to be fair for the Quora dataset which only had succinct questions. Properties of the datasets are summarized in Table~\ref{tab:1}, including total number of question pairs and Words Per Question (WPQ).

\begin{table}[ht!]
\scriptsize
\centering
\begin{tabular}{|c|c|c|c|c|}
 \hline
 Property & Quora & AskUbuntu & EnglishSE \\
 \hline
Question Pairs & 404,303 & 131,271 & 33,661 \\
 \hline
Max WPQ & 237 & 33 & 32 \\
 \hline
Mean WPQ & 11.0 & 8.7 & 8.9 \\
 \hline
\end{tabular}
\caption{Dataset Properties}
\label{tab:1}
\end{table}

\subsection{Data Preprocessing}
Each question was tokenized, and question pairs whose data types do not match were filtered. Non-English questions were removed by checking for non-English vowels. We also performed stop word removal, lemmatization, and stemming. Finally, abbreviated forms such as ``what's'' and ``i'm'' were transformed to their unabbreviated forms of ``what is'' and ``i am'' respectively.

\subsection{Deep Neural Network Models}
\begin{figure}[ht!]
\centering
\includegraphics[scale=0.53]{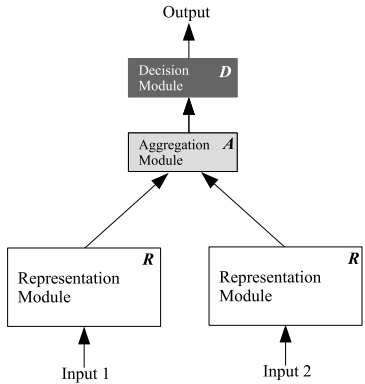}
\caption{Siamese Neural Network (SNN) for Duplicate Question Detection}
\label{fig:2}
\end{figure}

Underlying semantic similarity between questions can be learned with a better numerical representation of the texts, such as the ones learned through neural network models. The datasets we used have sufficient attributes to be used with a variety of deep neural network models. Siamese neural networks (SNN) have been popularly used to compare two objects and find similarity relationships between them~\cite{Chopra2005}. A salient feature of these Siamese networks is that they employ two sub-networks, which share parameters, thus reducing the number of parameters to learn, and give a consistent representation for the two objects being compared. We adapted a similar architecture to compare question pairs, and to determine whether they are duplicates. In Figure~\ref{fig:2} we illustrate an abstracted view of this adapted architecture to the duplicate question problem. Our outlined architecture features three major modules: i) Representation module ($R$), ii) Aggregation module ($A$), and iii) Decision module ($D$).

The representation module learns the representation of a question. This typically consists of an Embedding layer ($E$), and either Recurrent Neural Network (RNN) layers (e.g. Long Short Term Memory (LSTM) layers), or Convolution Neural Network (CNN) layers, and, optionally a few fully connected layers to flatten and summarize the output as a concise vector representation. The embedding layer converts the question words to vectors in the embedding space; we use the GloVe~\cite{Pennington2014} pre-trained word embeddings for this. 

The aggregation module takes the representations of the question pairs and performs an aggregation operation to prepare them for the decision module. $e^{-\mid Q_1 - Q_2 \mid}$ (negative absolute exponential distance) and the simple vector concatenation are two such successful aggregation methods we experimented with. 

The output of the aggregation module is fed to the decision module, which consists of one or more fully connected layers and a decision node with \texttt{sigmoid} activation at the end. We used \texttt{nadam} as the optimization function and binary cross entropy as the loss function. The datasets were split into 60\% for training, 20\% for validation, and 20\% for testing. Training and validation subsets were used to tune for the optimal hyper-parameter combinations, which included the optimal number of iterations to train, types and number of layers in the representation module, type of aggregation, number of nodes in each layer, and the best input length.
%

\subsection{Gradient Tree Boosting Models}
Gradient Tree Boosting (GTB) is a popular machine learning method which uses an ensemble of weak prediction models (typically decision trees) to build a strong predictor. It has shown strong results in various real-world applications~\cite{Chen2016}. Efficient gradient boosted tree implementations such as XGBoost have demonstrated very good performance in large datasets. Given the robustness of it, we experimented with a GTB based binary classification model to address the duplicate question detection problem.

As input features to the above GTB model, we used more than 40 hand-crafted features reflecting the semantic as well as structural similarities between two questions. These features included many traditional and non-traditional distance metrics such as TF-IDF distance, word movers distance, graph based structural question similarity distances, Word2Vec-based distances~\cite{Mikolov2013}, and Doc2Vec-based distances~\cite{Lau2016}.

\subsection{Transferability of Neural Networks}
Transfer learning (TL) aims to utilize the knowledge learned from a better performing source domain to increase the performance of an under-performing target domain with insufficient or sparsely labeled examples~\cite{Semwal2018}. Prior work in deep neural network-based computer vision models indicates that transfer learning can be successfully utilized~\cite{Shin2016}. Recently, NLP applications have used similar ideas to improve performance in certain target domains~\cite{Semwal2018, Mou2016}. 

In this work we explore the possibility of transferring and utilizing knowledge learned from large datasets such as Quora to improve the performance in other target domains such as Ask Ubuntu or English. Our intention is that this will lead to generally improved duplicate question detection across domains. Specifically, we adopt the INIT TL approach~\cite{Mou2016}, which uses parameters trained on a source domain to initialize parameters of the target domain's model. 

With INIT, we first trained a neural network model on the source dataset and experimented with three initialization states, $I_i$, on the target model: i) initialize the target parameters using source parameters but freeze further training, denoted $I_1$, ii) initialize the target parameters using source parameters and fine tune it further on target dataset, denoted $I_2$, and iii) random initialization, denoted $I_3$. We experimented with combinations of these initialization states on each module of our SNN ($R$epresentation [without the Embedding layer], $A$ggregation, and $D$ecision) and the $E$mbedding layer, and reported the best results obtained. Some example configurations are [$E(I_2)$, $R(I_3)$, $A(I_3)$, $D(I_3)$], [$E(I_2)$, $R(I_2)$, $A(I_3)$, $D(I_3)$], and [$E(I_2)$, $R(I_2)$, $A(I_2)$, $D(I_2)$].

\section{Results and Discussion}
\begin{figure}[ht!]
\centering
\includegraphics[scale=0.56]{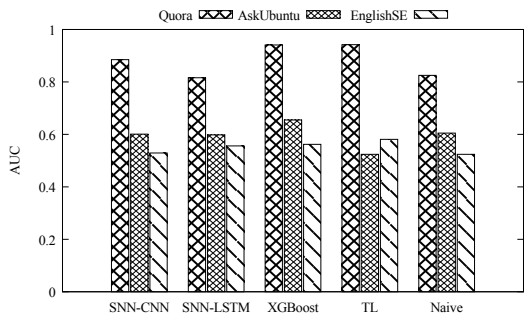}
\caption{Results from Various Machine Learning Approaches on Heterogeneous Datasets}
\label{fig:3}
\end{figure}

For performance evaluation, we use Area Under the Curve (AUC) metric. Our results are presented in Figure~\ref{fig:3}, including an additional na\"{\i}ve approach, in which we trained a model by combining training data from all three of our datasets. We then used the trained model to make predictions on the individual hold out test sets. We achieved state-of-the-art performance with Quora dataset using XGBoost, with 94.1\% AUC. This XGBoost approach was also best for the AskUbuntu dataset with 65.5\% AUC. For TL, we selected Quora as our source domain with neural networks as the preferred model, and models based on AskUbuntu and English SE as our targets. The TL approach gave the best performance for the English SE dataset at 58.1\% AUC, but only slightly better than XGBoost at 56.2\%. On the other hand Ask Ubuntu did not gain any improvement through transfer learning indicating the context specific difference in the duplicate question detection task.

For all the approaches, there were significant differences between performance across the datasets. While our approaches performed considerably well on the Quora dataset, the AskUbuntu and English SE datasets did not give comparatively good results even with the TL approach. This indicates that, across domains, the knowledge which can be positively transferred is low and the meaning of duplicates is vastly different.

Hence, the results provide some support for the alternative hypothesis that semantic representation of duplicates is significantly affected by specific domains. We postulate that the nature of the domain's language makes it difficult to predict duplicate text-pairs across domains. For example, Quora has simplified layperson English words, AskUbuntu has technical jargon and acronyms, while English SE has academic phrases.

\section{Related Literature}
We can classify the techniques for similar or duplicate question retrieval into three categories: i) translation models, ii) topic models, and iii) deep learning approaches. For methods using translation models, phrase-based translation models in community-based question retrieval have proven more effective because they seem to capture contextual information in modeling the translation of phrases as a whole, rather than translating single words in isolation~\cite{Zhou2011,Xue2008}. In addition, while these studies showed some promising results for detecting similar text-pairs, we did not gain any significant improvements in performance by using translation models on our duplicate text-pairs datasets. 

Some research has explored the use of topic models: first latent topics are identified for each question pair; and then four similarity scores are computed using their titles, descriptions, latent topics and tags~\cite{Zhang2015}. Others have proposed a supervised question-answer topic modeling approach, which assumes that questions and answers share some common latent topics~\cite{Zhang2014}.

Deep learning approaches have been investigated for identifying semantically equivalent questions as well as duplicate postings. Experiments have shown that a CNN can achieve high performance when word embeddings are pre-trained on in-domain data~\cite{Bogdanova2015}. SNN have also been used for similar question retrieval, where the network learns the similarity metric for question pairs by leveraging question-answer pairs available in CQA archives~\cite{Das2016}. The Siamese architecture approach has also been used for duplicate pairs classification specifically on the Quora data~\cite{Homma2016}. Our study exploits that research, and explores the general-purpose application of various machine learning approaches to heterogeneous datasets, including experimenting with TL~\cite{Semwal2018} to adapt for different domains.

\section{Conclusion}
We presented the results of our ongoing research work on duplicate text-pair detection in community question answering, using a variety of machine learning approaches. Our goal was to determine if a robust pipeline or model could be built that can predict duplicate question pairs across heterogeneous datasets. Additionally, we tested the null hypothesis that the meaning of ``duplicate'' is universal across domains. Our selected learning methods included deep neural networks, gradient tree boosting, and transfer learning. We demonstrated state-of-the-art results for accurately predicting text-pairs on Quora. Our empirical analysis supports the alternative hypothesis on the meaning of duplicates. For future work, we intend to investigate more complex configurations for transfer learning approaches using a larger collection of domain-diverse Q\&A datasets while tackling the data imbalance problem.

\balance

\nocite{*}
\bibliography{acl2018}
\bibliographystyle{acl_natbib}

\end{document}